\documentclass[runningheads]{llncs}

\usepackage{amsmath,epsfig}
\usepackage{multirow}
\usepackage{hyperref}
\usepackage{amssymb}
\usepackage{color}
\usepackage{caption}
\usepackage[cmyk]{xcolor}
\usepackage{bm}
\hypersetup{
    colorlinks,
    citecolor=cyan,
    linkcolor=red, 
    urlcolor=green 
}

\usepackage[T1]{fontenc}
\usepackage{graphicx}
\begin{document}
%
\title{Enhancing Table Structure Recognition via Bounding Box Guidance
}

%
%
\author{Lei Hu \and
Shuangping Huang\thanks{Corresponding author.}}
\authorrunning{L. Hu et al.}
\titlerunning{BGTR}
%
\institute{South China University of Technology, Guangzhou, China \email{eehulei@mail.scut.edu.cn, eehsp@scut.edu.cn} \\}
\maketitle              
\begin{abstract}
Table Structure Recognition (TSR) aims to extract the bounding boxes of cells and table structure (e.g., HTML) from table images. Although current approaches have made significant progress, the latest image-to-sequence methods overlook the explicit utilization of the bounding box information when predicting HTML sequences, leading to error predictions in complex scenes. In this paper, we introduce a novel framework {\bf BGTR} ({\bf B}ounding Box-{\bf G}uided {\bf T}able {\bf R}ecognizer). To more effectively utilize bounding box information, we first predict the bounding boxes of cells and then use this information to guide the generation of HTML sequences. While utilizing bounding box information can enhance the accuracy of HTML sequences, for natural scene tables, the data volume is too small to allow for sufficient training of bbox-guided HTML generation. In response, we adopt a progressive training method for natural scene tables and introduce {\bf SNSTab}, a synthetically generated natural scene table dataset. Our experiments on five benchmark datasets demonstrate SOTA performance.

\end{abstract}

\section{Introduction}
\label{sec:intro}

\begin{figure}[t]
    \centering
    \begin{minipage}{\columnwidth}
        \centering
        \includegraphics[width=0.7\columnwidth]{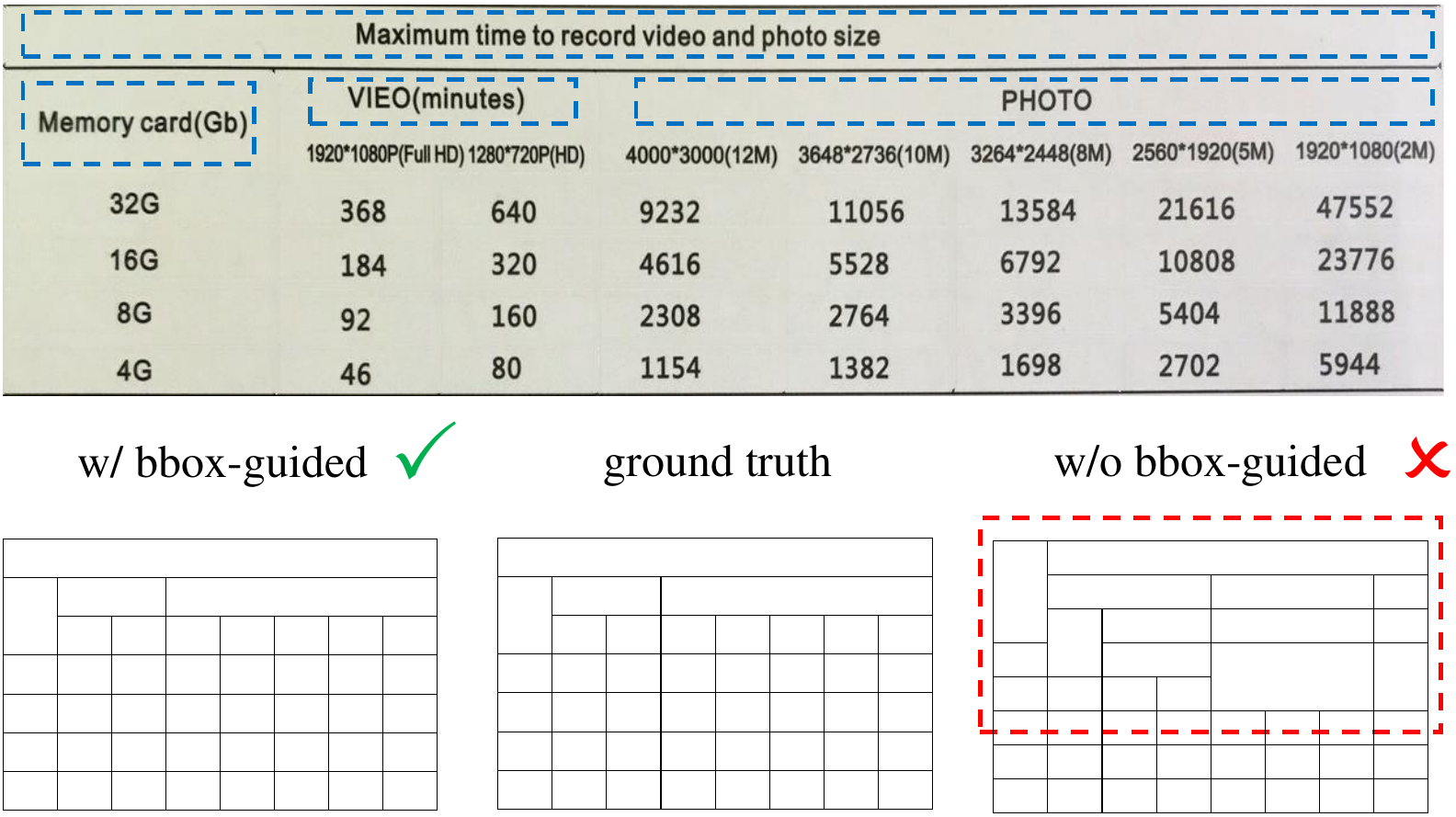}
        \captionsetup{labelformat=empty}
        \caption*{(a)}
    \end{minipage}
    
    \vspace{5pt}
    \begin{minipage}{\columnwidth}
        \centering
        \tabcolsep=0.2cm
        \renewcommand\arraystretch{1.1}
        \begin{tabular}{cccc}
            \hline
            \multicolumn{4}{c}{\textbf{Digital Document Tables}} \\ \hline
            Dataset & Samples & Datasets & Samples \\ \hline
            PubTabNet~\cite{EDD} & \textcolor{green}{500K} & FinTabNet~\cite{GTE} & \textcolor{green}{100K} \\
            SynthTabNet~\cite{Tableformer} & \textcolor{green}{600K} & PubTables-1M~\cite{pubtables} & \textcolor{green}{758K} \\ \hline
            \multicolumn{4}{c}{\textbf{Natural Scene Tables}} \\ \hline
             TabRecSet~\cite{TabRecSet} & \textcolor{red}{38K} & WTW~\cite{WTW} & \textcolor{red}{14K} \\
             iFLYTAB~\cite{SEMV2} & \textcolor{red}{17K} & TAL~\cite{TAL} & \textcolor{red}{15K}  \\ \hline
        \end{tabular}
        \captionsetup{labelformat=empty}
        \caption*{(b)}
    \end{minipage}

    \caption{The motivation behind the proposed method. (a) Comparison of HTML visualization with and without \textbf{bbox-guided} generation on TabRecSet~\cite{TabRecSet}, the \textcolor{red}{red} dotted boxes indicate error results, the \textcolor{cyan}{blue} dotted boxes indicate spanning cells. The results indicate that using bboxes for guidance achieves better performance in spanning cells. (b) A comparison between two types of datasets reveals that the sample size of the \textbf{digital document} table dataset is significantly larger than that of the \textbf{natural scene} table dataset.}
    \label{fig:1}
\end{figure}

Tables are a crucial medium for structured information dissemination. Table detection (TD) aims to extract the position of tables from document images, and many methods~\cite{Cdec-net,gemelli2022graph,shehzadi2023towards} have shown excellent results. Table Structure Recognition (TSR) aims to transform images containing tables into structured data, which is both crucial and challenging. Leveraging the advancements of transformer~\cite{transformer}, which have proven highly effective in various fields~\cite{dai2023disentangling}, image-to-sequence methods~\cite{Tableformer,VAST,TableMaster,chen2022complex} have demonstrated promising results in TSR. These methods employ an encoder-decoder architecture to simultaneously predict the HTML (Hyper Text Markup Language) sequence and the bounding box (bbox) of table cells. In predicting HTML sequences, they rely solely on image information and overlook the explicit utilization of bbox information. However, table structure and formatting can be highly complex, and bbox information is essential for parsing the structure of tables. Therefore, exclusive reliance on image information may lead to error predictions in complex scenes, like spanning cells (Fig.~\ref{fig:1} (a)). In this paper, we introduce a novel framework {\bf BGTR} ({\bf B}ounding Box-{\bf G}uided {\bf T}able {\bf R}ecognizer). Unlike previous image-to-sequence methods~\cite{Tableformer,VAST,TableMaster,chen2022complex}, we explicitly utilize bbox information to obtain accurate HTML sequences. We first use a Bbox Predictor to predict bboxes. Then, during HTML sequence decoding, we enable the Bbox-Guided Structure Decoder to perceive both the image and bbox information of the table, resulting in accurate HTML sequences.

Although utilizing bbox information can improve the accuracy of the HTML sequence, for natural scene tables, as shown in Fig.~\ref{fig:1} (b), the data volume is small, and the structure and style of tables in natural scenes are complex, making it insufficient for adequate training of bbox-guided HTML generation. In response, we adopt a progressive training method for natural scene tables and introduce {\bf SNSTab}. Progressive training method includes a foundation training stage and a advancement training stage. In the foundation training stage, we aim to train the model with a large number of tables from natural scenes, thereby enabling it to learn how to more effectively utilize bbox information for guiding the generation of HTML sequences, this approach leads to improve the model's foundational understanding of tables. In the advancement training stage, training is conducted on a specific natural scene table dataset (e.g., TabRecSet~\cite{TabRecSet} and iFLYTAB~\cite{SEMV2}). {\bf SNSTab} is a synthetically generated natural scene table dataset containing 500k table images for the foundation training stage. It includes wired tables, wireless tables, inclined tables, and curved tables, featuring diverse table structures and backgrounds. This variety enables the model to comprehensively learn various aspects of table knowledge during the foundation training stage, thereby achieving better results in complex scenes like spanning cells and  deformed tables, as shown in Fig.~\ref{fig:4}.

Extensive experiments demonstrate the effectiveness of our proposed BGTR and the progressive training method, achieving state-of-the-art performance on five public benchmarks.

To sum up, our contributions are as follows:

\begin{itemize}
  \item We propose {\bf BGTR}, a novel framework that explicitly utilizes bbox information for guiding HTML sequence generation, which aims at enhancing structural recognition accuracy in challenging table scenes. 

  \item To ensure that bbox-guided HTML generation is adequately trained in natural scenes, we adopt a progressive training method and introduce {\bf SNSTab}, a synthetically generated natural scene table dataset for the foundation training stage. 

  \item Our experiments on five benchmark datasets demonstrate state-of-the-art performance.
\end{itemize}

\section{Related Work}

\subsection{ Table Structure Recognition}

With the rapid development of deep learning, a variety of table structure recognition methods have emerged, which can be divided into three categories: graph-based methods, split-and-merge methods, and image-to-sequence methods.

{\bf Graph-based Methods.} These methods utilize cells or text boxes as the basic elements of the table, employing a graph network to determine the row and column relationships between them. GraphTSR~\cite{chi2019complicated} utilized graph attention networks to the TSR task, determining the row and column relationships of adjacent cells through graph edge classification. TabStruct-Net~\cite{raja2020table} implemented a unified end-to-end framework for cell detection and cell relationship analysis. GFTE~\cite{GFTE} employed a graph-based convolutional network that integrates image features, position features, and textual features to predict relationships between cells. NCGM~\cite{NCGM} enabled cooperation among geometry, appearance, and content modalities, leveraging their interaction to enhance multi-modal representation in intricate situations. However, these methods are limited by their reliance on additional bbox data or OCR accuracy, leading to potential errors in table structure recognition, and additionally, they need complex post-processing methods.

{\bf Split-and-merge Methods.} Typically, these methods comprise two models: the split model and the merge model. The split model initially detects the row and column regions of the table and then intersects them to obtain the grid cells of the table. Subsequently, the merge model is employed to determine which adjacent grid cells need to be merged. SPLERGE~\cite{tensmeyer2019deep} became the first to use the split-and-merge framework for the TSR task, addressing an issue where previous methods struggled with resolving spanning cells. By utilizing textual information, SEM~\cite{SEM} achieved enhanced results on complex tables with spanning cells. To address geometric distortion in table images, TSRFormer~\cite{Tsrformer} approached the detection of row and column regions as a linear regression problem. However, two-stage training can be complex and resource-intensive, potentially leading to longer training times and difficulties in optimization compared to more streamlined, end-to-end methods.


{\bf Image-to-sequence Methods.} These methods treat the table as a structured sequence (e.g., HTML or LATEX), using an encoder-decoder framework to convert the table image into a structured sequence that fully describes the table structure. EDD~\cite{EDD} employed a CNN-based encoder to extract the visual features from table images and utilized two LSTM-based decoders to simultaneously recognize the table structure and cell content. TableMaster~\cite{TableMaster} introduced a transformer-based~\cite{transformer} architecture, achieving significant progress in the TSR task by recognizing the table structure and cell bboxes simultaneously. Based on TableMaster~\cite{TableMaster}, VAST~\cite{VAST} treated bbox prediction as a coordinate sequence generation task and introduced a visual-alignment loss that significantly improved bbox accuracy. However, bbox information is essential for parsing the structure of table, unlike previous methods that produce inaccurate HTML sequence predictions in complex table scenes due to the lack of bbox information, this paper utilizes bbox information to guide the generation of HTML sequences, resulting in more accurate HTML sequences.

\subsection{Existing Datasets}

While the size of table datasets has significantly increased, existing datasets primarily focus on digital documents~\cite{Tableformer,EDD,GTE,pubtables}, such as PDF files. Building a digital document table dataset is relatively straightforward because annotated information can be directly extracted from PDF files. However, tables captured in natural scenes through cameras cannot be automatically annotated, and manual annotation is a time-consuming process. Additionally, natural scene tables are more complex, often inclined, rotated, and curved, further increasing the annotation difficulty. Due to these challenges, there is a substantial disparity in the number of table datasets between natural scenes and digital documents, as illustrated in Fig.~\ref{fig:1} (b). To address this issue, we propose a large-scale synthetically generated natural scene table dataset.

\begin{figure}[t]
\centerline{\epsfig{figure=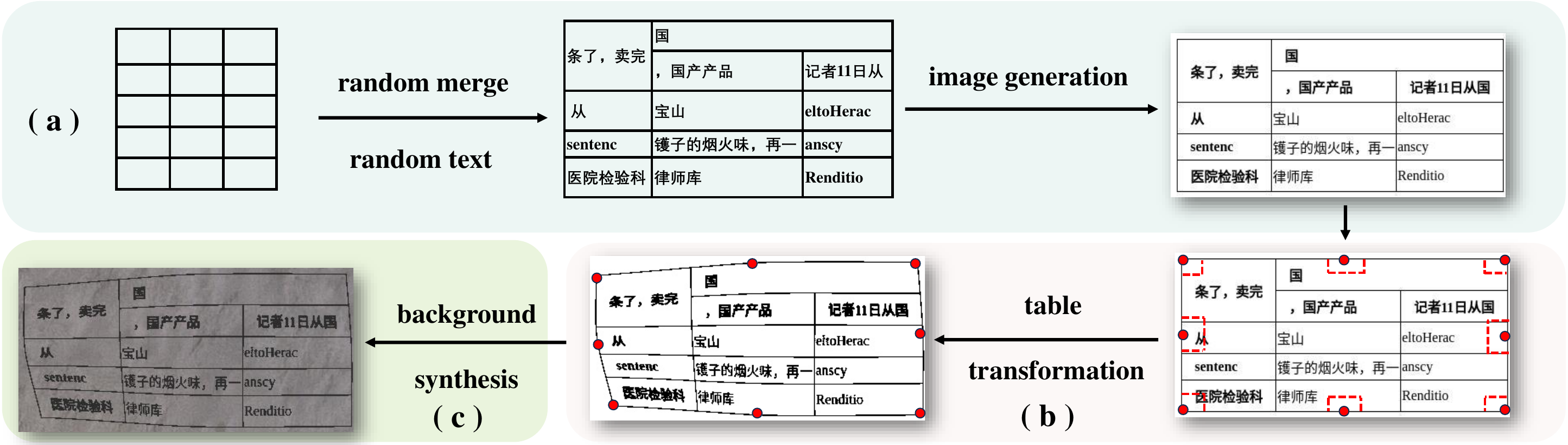,width=12cm}}
\caption{The production process of SNSTab: (a) table generation. (b) table transformation. (c) background synthesis.}
\label{fig:9}
\end{figure}

\section{SNSTab}
\label{SNSTAB}

SNSTab contains 500k synthetic images of natural scene tables, including wired tables, wireless tables, provincial line tables, inclined tables, curved tables and large tables. Although image generation has achieved significant success in other fields~\cite{one-dm2024}, its application in table recognition remains quite limited. To our knowledge, SNSTab is the {\bf first} large-scale natural scene table synthesis dataset. SNSTab's annotations contain the coordinates of the table cells, the text inside the cells, and the HTML sequence that describes the table structure. The creation of the SNSTab dataset involves three phases: table generation, table transformation, and background synthesis, as shown in Fig.~\ref{fig:9}.

{\bf Table generation.} This step is to generate digital document table images. We randomly generate table images based on the open source tool Table Generation\footnote{https://github.com/WenmuZhou/TableGeneration}. First, we will generate a grid with a random number of rows and columns; Then, we will randomly merge the adjacent grids to get spanning cells, and generate random text for each grid; Finally, we convert the above table into HTML sequences, and get the final table image through the browser rendering.

{\bf Table transformation.} Since tables in the nature scene tend to be inclined or rotated. Therefore, after automatically generating tables, we apply thin plate spline (TPS)~\cite{TPS} to randomly transform them. This simulation captures the complexities observed in natural scenes. As shown in Fig.~\ref{fig:9} (b), we take the four vertices of the table image and the midpoints of the four sides as the source points, the target points are then obtained by randomly moving the source points within the range of the red dotted line. After the TPS transformation, the coordinates of the cells are also transformed.

{\bf Background synthesis.} In addition to their complex structures, tables in natural scenes often exhibit a variety of backgrounds. We captured 400 background images of natural scenes, including paper, walls, daily-life items, and more. For each table image, a random background image is first selected, and then a random area of the same size as the table image is extracted from the background image. Finally, the table image is merged with the selected background to produce the final image.

For more details of the dataset and for additional dataset samples, please refer to the supplementary materials.

\begin{figure}[t]
\centerline{\epsfig{figure=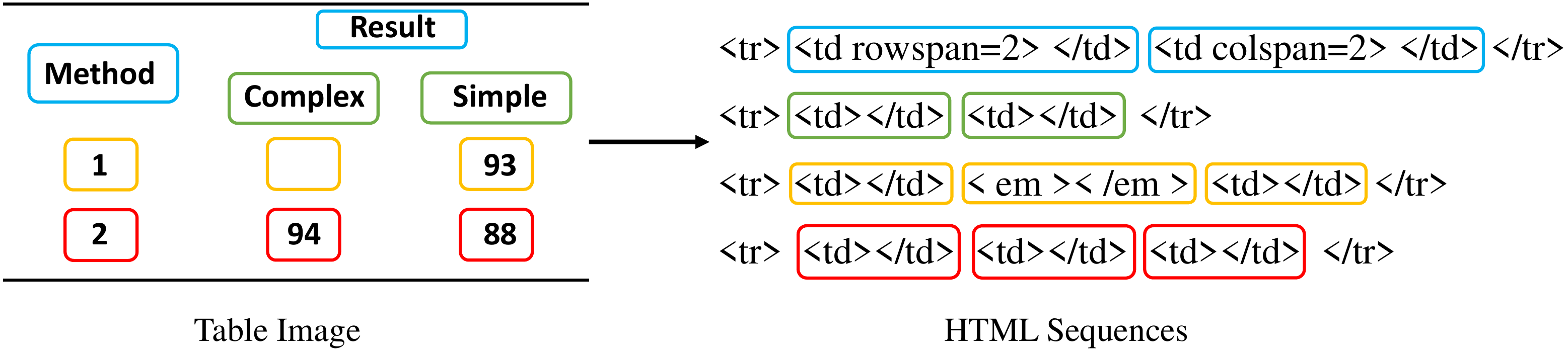,width=12cm}}
\caption{A simple example of using HTML sequences to represent a table structure, different colors in the figure represent different rows of the table.}
\label{fig:8}
\end{figure}

\section{Method}

\subsection{Preliminary}

In this paper, we utilize HTML sequences to represent the table structure, as shown in Fig.~\ref{fig:8}. Given a table image, our model outputs HTML sequences of the table and the corresponding cell bboxes.  To better facilitate prediction, we tokenize HTML sequences into HTML tokens. For cells without spanning, non-empty cells and empty cells are denoted by $<td></td>$ and $<em></em>$, respectively. In the case of spanning cells, the tokens are divided into three parts: $<td$, $colspan=N$ or $rowspan=N$, and $></td>$. Here, $<td$ indicates the beginning of the spanning cells, $N$ specifies the count of cells that are spanning, and $></td>$ marks the end of the spanning cells. $<tr>$ and $</tr>$ respectively represent the beginning and the end of each row in a table. We use ${H}_{N}={\{{h}_{i}}\}_{i=1}^{N} \in \mathbb{R}^{N \times 1}$ to denote HTML sequences, where N is the sequence length and ${h}_{i}$ denotes the $i$-th HTML token. We use ${B}_{N}={\{{b}_{j}}\}_{j=1}^{N} \in \mathbb{R}^{N \times 4} $ to denote the bbox of table cells. For each cell, its bbox is represented as $[{x}_{1}, {y}_{1}, {x}_{2}, {y}_{2}]$, where $[{x}_{1}, {y}_{1}]$ denotes the coordinates of the top-left corner, and $[{x}_{2},{y}_{2}]$ represents the coordinates of the bottom-right corner. Moreover, the HTML tokens have a one-to-one correspondence with the bboxes, and the bbox value is non-zero only if the HTML token is $<td></td>$ and $<td$.

\subsection{Overall Architecture}

\begin{figure}[t]
\centerline{\epsfig{figure=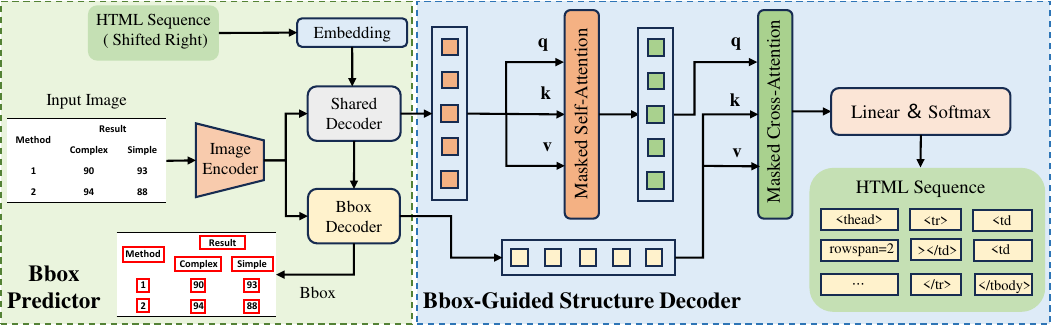,width=12cm}}
\caption{Architecture of  BGTR. The Bbox Predictor aims to acquire the bboxes of table cells and consists of three parts: an image encoder, a shared decoder, and a bbox decoder. The Bbox-Guided Structure Decoder generates the HTML sequence.}
\label{fig:2}
\end{figure}

The overall framework of BGTR is illustrated in Fig.~\ref{fig:2}. Given a table image, denoted as $ \bold P \in \mathbb{R}^{H\times W\times3}$, where H and W represent the height and width of the image, respectively. We employ an Image Encoder to extract image features, resulting in the feature map $ \bold {F}_{encoder} \in \mathbb{R}^{\frac{H}{8}\times \frac{W}{8}\times d}$, where $d$ denotes the dimension of the features. After applying 2D positional encoding, the flattened image features are obtained as $ \bold {F}_{image} \in \mathbb{R}^{\frac{HW}{64}\times d}$. The image features $\bold {F}_{image}$ are further fed into the Shared Decoder for decoding, resulting in decoded features $ \bold {F}_{share}^{1} \in \mathbb{R}^{N \times d} $. The Shared Decoder is used to reduce the gap between the image and the sequence, making it more aligned with the sequential features. $\bold {F}_{share}^{1}$ is first fed into the Bbox Decoder to obtain bboxes ${B}_{N}$. Then, $\bold {F}_{share}^{1}$ is sent into the Bbox-Guided Structure Decoder (Sec.~\ref{subsec:Bbox-Guided Structure Decoder}), using the predicted bbox information to guide the generation of HTML sequences ${H}_{N}$. For additional details regarding the Image Encoder, Shared Decoder, and Bbox Decoder, please refer to Sec.~\ref{subsec:details}.

\subsection{Bbox-Guided Structure Decoder}
\label{subsec:Bbox-Guided Structure Decoder}

To more effectively utilize bbox information, we first predict the bboxes of cells and then utilize the bbox information to enhance the accuracy of HTML sequence prediction. Since we employ an autoregressive decoding approach, we utilize parallel training methods during training to accelerate the training speed.   Specifically, the Bbox-Guided Structure Decoder receives $ \bold {F}_{share}^{1} \in \mathbb{R}^{N \times d}$ from the Shared Decoder and $ \bold {F}_{bbox} \in \mathbb{R}^{N \times d}$ from the Bbox Decoder as input. In the Bbox Decoder, after $ \bold {F}_{bbox}$ passes through a linear layer and a sigmoid layer, the bboxes ${B}_{N} \in \mathbb{R}^{N \times 4}$ are obtained. $ \bold {F}_{share}^{1}$ initially passes through a masked self-attention layer, resulting in $ \bold {F}_{share}^{2} \in \mathbb{R}^{N \times d}$. Here, the mask refers to the prediction of the current time step HTML token being based on the output of previous time steps. Subsequently, $ \bold {F}_{share}^{2}$ and $ \bold {F}_{bbox}$ are fed into a masked cross-attention layer. Here, the mask indicates that the prediction of the current time step HTML token is based on the bbox outputs of both the current and previous time steps. $ \bold {F}_{share}^{2}$ serves as the query vector, while $ \bold {F}_{bbox}$ serve as the key/value vectors. By utilizing the cross-attention mechanism, bbox information $ \bold {F}_{bbox}$ becomes effectively integrated into $ \bold {F}_{share}^{1}$. The use of $ \bold {F}_{bbox}$ for decoding allows the model to comprehensively understand the position and relative relationships of each cell while predicting HTML sequences. This process allows the model to generate HTML sequences guided by bbox information. After passing through a linear layer and a softmax layer, the decoder's output yields the final HTML sequences ${H}_{N}={\{{h}_{i}}\}_{i=1}^{N} \in \mathbb{R}^{N \times 1}$.

\begin{figure}[t]
\centerline{\epsfig{figure=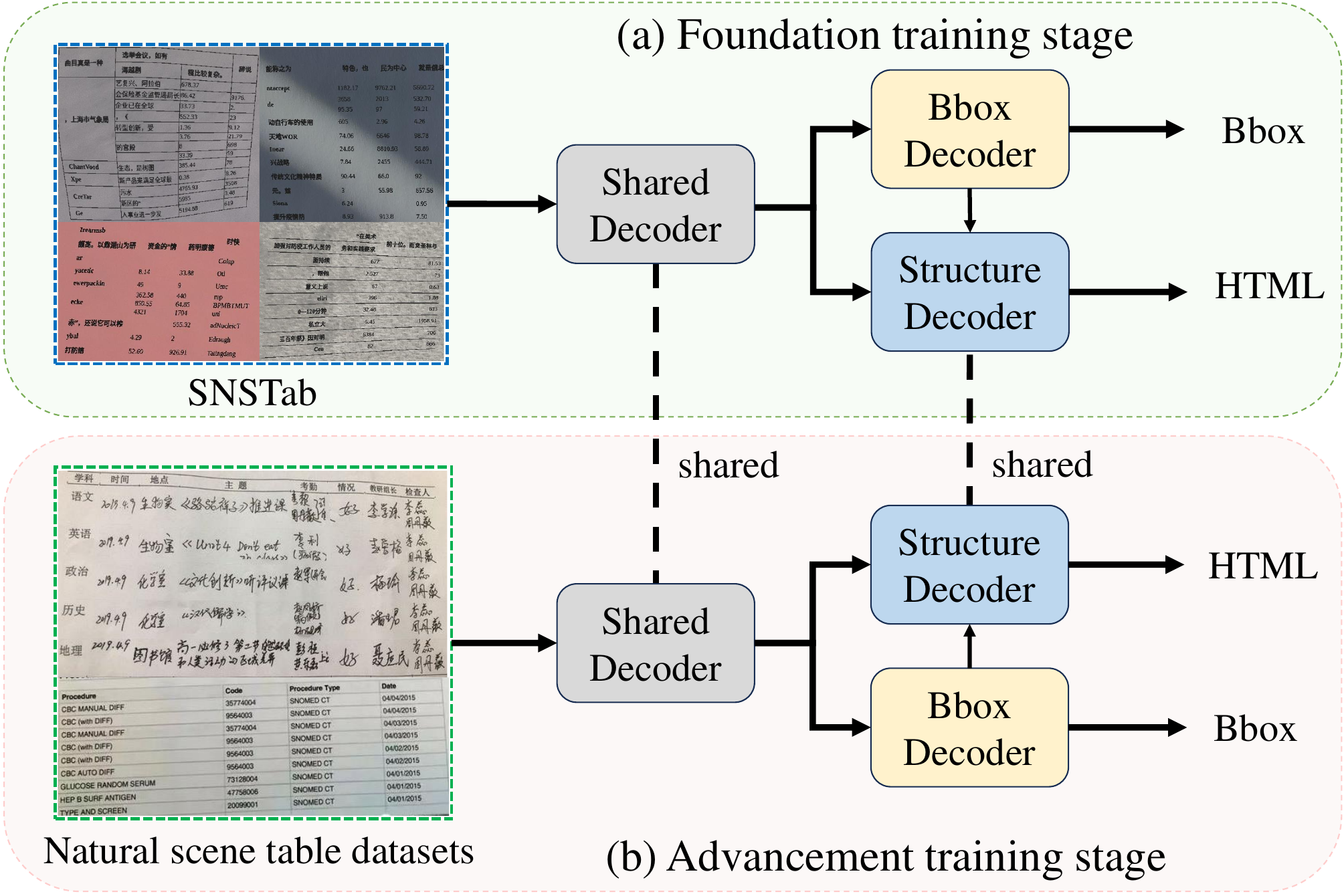,width=9cm}}
\caption{An overview of the progressive training method. (a) Foundation training stage training on SNSTab. (b) Advancement training stage training on natural scene table datasets (e.g., TabRecSet~\cite{TabRecSet} and iFLYTAB~\cite{SEMV2}).}
\label{fig:3}
\end{figure}

\subsection{Progressive Training Method}

As illustrated in Fig.~\ref{fig:3}, this section will discuss the implementation of the progressive training method.

{\bf Foundation training stage.} In the foundation training stage, the purpose is for the model to acquire common knowledge about tables. Due to the diverse types and varied structures of tables in natural scenes, the foundation training stage requires a large number of data samples. Based on this, we introduce SNSTab, a large synthetic table dataset in natural scenes. For further details about SNSTab, please refer to Sec.~\ref{SNSTAB} and the supplementary material. After completing the foundation training stage on SNSTab, the model develops a foundational capability for recognizing table structures in natural scenes. Additionally, it can learn how to effectively utilize bbox information to guide the generation of HTML sequences, particularly in complex scenes such as spanning cells and deformed tables (Fig.~\ref{fig:4}).

{\bf Advancement training stage.} Building on the foundation training stage, the advancement training stage is conducted on a specific natural scene table dataset. With the common knowledge acquired in the foundation training stage, the model demonstrates improved convergence speed and enhanced overall training effectiveness in the advancement training stage. And in this stage, the Shared Decoder and the Bbox-Guided Structure Decoder are initialized using the training from the foundation training stage. Due to certain differences in the data between the two stages, the Bbox decoder is trained from scratch.

Through the progressive training process, the issue of insufficient data leading to inadequate training of bbox-guided HTML generation in natural scenes has been significantly alleviated.

\subsection{Loss Functions}

Our model adopts an end-to-end training approach and includes two loss functions. For the Bbox Decoder, ${L}_{1}$ loss is employed to supervise the prediction of bboxes, which is denoted as $\mathcal{L}_{bbox}$. For Bbox-Guided Structure Decoder, cross-entropy loss is utilized to supervise the prediction of HTML tokens, which is denoted as $\mathcal{L}_{html}$. The final loss function is formulated as follows:
\begin{eqnarray}
\mathcal{L}={\lambda }\mathcal{L}_{html}+\mathcal{L}_{bbox} \text{,}
\end{eqnarray}
where ${\lambda }$ is the hyperparameter.

\section{Experiments}

\subsection{Datasets and Evaluation Metric}

\subsubsection{Datasets.} Our method is evaluated on five popular public benchmarks, including TabRecSet~\cite{TabRecSet}, iFLYTAB~\cite{SEMV2}, PubTabNet~\cite{EDD}, FinTabNet~\cite{GTE} and SynthTabNet~\cite{Tableformer}.

{\bf TabRecSet}~\cite{TabRecSet} is a natural scene table dataset featuring tables from diverse scenes with various forms. It has 32.07K images and 38.17K tables, the number of images is not equal to the number of tables because some images contain multiple tables. As TabRecSet did not provide a predefined split, we randomly divided the dataset into train and test splits(80\%,20\%), resulting in 30.6k training table images and 7.5k testing table images. 

{\bf iFLYTAB}~\cite{SEMV2} has 12,104 training samples and 5,187 testing samples. It contains both wired and wireless tables from natural scenes and digital documents.

{\bf PubTabNet}~\cite{EDD} contains 500,777 training images and 9,115
validating images, each accompanied by annotation information detailing the table structure and text content along with their positions.  All the tables are extracted from the scientific articles, and annotations are automatically obtained from the PDF source files.

{\bf FinTabNet}~\cite{GTE} is a large-scale dataset containing 91596 training tables, 10,635 validating tables and 10,656 testing tables. All the tables are sourced from the annual reports of the S\&P 500 companies. Following~\cite{GTE,Tableformer,VAST,gridformer}, we use validating sets for testing.

{\bf SynthTabNet}~\cite{Tableformer} is a synthetically generated dataset with diverse table styles, complex structures, and an increased number of rows and columns. It contains 480k training images, 60k validating images, and 60k testing images. In addition to the bounding boxes of the non-empty cell, it also has the bounding boxes of the empty cell.

\subsubsection{Evaluation Metric.} The Tree-Edit-Distance-based Similarity (TEDS)~\cite{EDD} is employed as the evaluation metric, treating tables as tree structures. To mitigate the impact of OCR errors on the final score, we also utilize TEDS-S to assess the accuracy of the table structure without the table content.

\begin{table}[t]
\begin{center}
\tabcolsep=0.05cm
\renewcommand\arraystretch{1.15}
\caption{Comparison with state-of-the-art methods. PT indicates that progressive training method is used. S indicates simple tables. C indicates complex tables. \textbf{Bold} indicates the best performance, while \underline{underline} indicates the second-best performance. $\bm{\star}$ indicates the image-to-sequence method. $\bm{\dagger}$ means pre-training on PubTabNet~\cite{EDD}.}
\vspace{2mm}
\label{tab:2}
\begin{description}
   \leavevmode \\ 
\end{description}
\begin{tabular}{lcccccccccc}
\hline
\multicolumn{1}{l}{\multirow{3}{*}{Method}} &  \multicolumn{2}{c}{PubTab} & \multicolumn{1}{c}{FinTab} & \multicolumn{1}{c}{SynthTab} & \multicolumn{3}{c}{TabRecSet} & \multicolumn{3}{c}{iFLYTAB} \\ \cline{2-11} 
\multicolumn{1}{c}{} & \multicolumn{1}{c}{\multirow{2}{*}{TEDS-S}} & \multicolumn{1}{c}{\multirow{2}{*}{TEDS}} & \multicolumn{1}{c}{\multirow{2}{*}{TEDS-S}} & \multicolumn{1}{c}{\multirow{2}{*}{TEDS-S}} & \multicolumn{3}{c}{TEDS-S} & \multicolumn{1}{c}{\multirow{2}{*}{TEDS-S}} \\ \cline{6-8} 
\multicolumn{1}{c}{} &  \multicolumn{1}{c}{} & \multicolumn{1}{c}{} & \multicolumn{1}{c}{} & \multicolumn{1}{c}{} & \multicolumn{1}{c}{S} & \multicolumn{1}{c}{C} & \multicolumn{1}{c}{All}  \\ \hline
 EDD $\bm{\star}$~\cite{EDD}  &  89.90 & 88.30 & 90.06 & - & 95.01 & 77.71 & 91.03$\bm{\dagger}$ & -  \\
 GTE~\cite{GTE}  &  93.01 & - & 87.10 & - & - & - & - & -  \\
 TableMaster $\bm{\star}$~\cite{TableMaster} &  96.04 & 96.16 & - & - & 97.20 & 84.11 & 94.14 &  84.63 \\
 SEM~\cite{SEM} &   - & 93.70 & - & - & - &-& - & 75.90 \\
 NCGM~\cite{NCGM} & - & 95.40 & - & - & - & - & - & -  \\
 TableFormer $\bm{\star}$~\cite{Tableformer} &  96.75 & 93.60 & 96.80 & \underline{96.70} & - & - & - & -   \\
 VAST $\bm{\star}$~\cite{VAST} &  97.23 & \underline{96.31} & \underline{98.63} & - & - & - & - & -  \\
 GridFormer~\cite{gridformer} &  97.00 & 95.84 & \underline{98.63} & - & - & - & - & -  \\ 
 
SEMv2~\cite{SEMV2} &  \underline{97.50} & - & - & - & - & - & - & \textbf{92.00} \\
 
TSRFormer~\cite{Tsrformer} &  \underline{97.50} & - & - & - & - & - & - & -  \\
 
 \hline
 BGTR $\bm{\star}$ &  \textbf{97.63} & \textbf{96.57} & \textbf{98.89} & \textbf{99.11} & \underline{98.35} & \underline{89.27} & \underline{96.23} & \underline{91.02}  \\ 

 BGTR (PT) $\bm{\star}$ &  - & - & - & - & \textbf{98.65} & \textbf{92.47} & \textbf{97.21} & \textbf{92.00}  \\ \hline
\end{tabular}
\end{center}
\end{table}

\begin{table}[t]
\vspace{-3mm}
\begin{center}
\tabcolsep=0.5cm
\renewcommand\arraystretch{1.0}
\caption{Comparison of cell bbox detection results on PubTabNet. PP indicates the post-processing.}
\vspace{2mm}
\label{tab:6}
\begin{tabular}{ccc}
\hline
Method &  mAP & mAP(PP) \\ \hline
EDD $+$ BBox~\cite{Tableformer} & 79.2 &  82.7 \\
TableFormer~\cite{Tableformer} & 82.1 & 86.8  \\
\textbf{BGTR} & \textbf{91.9} & -  \\ \hline
\end{tabular}
\vspace{-1mm}
\end{center}
\end{table}

\subsection{Implementation Details}
\label{subsec:details}

In this paper, the experimental settings are as follows: the table images are resized to $480 \times 480$, and the flattened image sequence length is 3600. The dimension of the features $d$ is 512. The multi-head number is 8. The maximum HTML sequence length is 500. We used Ranger~\cite{Ranger} as the optimizer, the mini-batch size is set to 8. For TabRecSet, PubTabNet, FinTabNet and SynthTabNet, we trained 25 epochs, the initial learning rate is established at 1e-3, and divided by 10 at 17 and 22 epochs. For iFLYTAB, we trained 120 epochs, the initial learning rate is established at 1e-3, and divided by 10 at 75 and 105 epochs. For the foundation training stage on SNSTab, we trained 3 epochs, the initial learning rate is established at 1e-3. Experiments are conducted using 2 NVIDIA GeForce RTX 3090 GPUs with 24GB of RAM memory.

We use the ResNet-50~\cite{resnet} combined with the Multi-Aspect GCA~\cite{master} module and 2D positional encoding to form the Image Encoder. To enhance the model's understanding of the 2D topology of table images, we employ 2D positional encoding to encode image features. The Shared Decoder comprises two identical stacked transformer~\cite{transformer,replaycad,immoe} decoding layers. The Bbox Decoder comprises a single transformer~\cite{transformer} decoding layer. The Bbox-Guided Structure Decoder comprises two identical stacked transformer~\cite{transformer} decoding layers.

\subsection{Comparison with Previous State-of-the-arts}

As shown in Table~\ref{tab:2}, our method not only outperforms non-image-to-sequence methods, but also outperforms the best image-to-sequence method.

{\bf Results on natural scene tables.} We evaluate the performance of our model on two natural scene table datasets: TabRecSet~\cite{TabRecSet} and iFLYTAB~\cite{SEMV2}. Given the absence of a baseline method in TabRecSet, TableMaster~\cite{TableMaster} is adopted as the baseline. We divide the dataset into two categories: simple (S) and complex (C). A table is considered complex if it contains spanning cells, otherwise, it is classified as a simple table. On TabRecSet, a TEDS-S score of 98.65\% for simple tables and 92.47\% for complex tables is achieved. Compared with baseline TableMaster, our method demonstrates improvements of 1.45\% on simple tables, 8.36\% on complex tables, and 3.07\% overall. On iFLYTAB, a TEDS-S accuracy of 92.00\% is achieved by our method, comparable to SEMv2~\cite{SEMV2} and outperforms other methods.

{\bf Results on digital document tables.} The performance of our model is also evaluated on three digital document table datasets: PubTabNet~\cite{EDD}, FinTabNet~\cite{GTE} and SynthTabNet~\cite{Tableformer}. For PubTabNet, similar to previous methods~\cite{VAST,gridformer,TRUST}, the OCR results are from the text detection method PSENet~\cite{PSENET} and text recognition method MASTER~\cite{master}, and we match the text bboxes to the cell bboxes as described in~\cite{TableMaster}. A TEDS-S score of 97.63\% and a TEDS score of 96.57\% are achieved on PubTabNet which outperforms other methods. For FinTabNet and SynthTabNet, TEDS-S scores of 98.89\% and 99.11\% are achieved, respectively. Compared with TableFormer~\cite{Tableformer}, our method exhibits improvements of 2.09\% and 2.41\% on FinTabNet and SynthTabNet, respectively.

In addition, we evaluate the performance of cell bbox detection on PubTabNet~\cite{EDD} using the PASCAL VOC mAP metric. As shown in Table~\ref{tab:6}, our method outperforms TableFormer~\cite{Tableformer} by 5.1\% even without using post-processing.

The results on five datasets validate the effectiveness of using bbox to guide the generation of HTML sequences.

\subsection{Visualization}

\begin{figure}[]
\centerline{\epsfig{figure=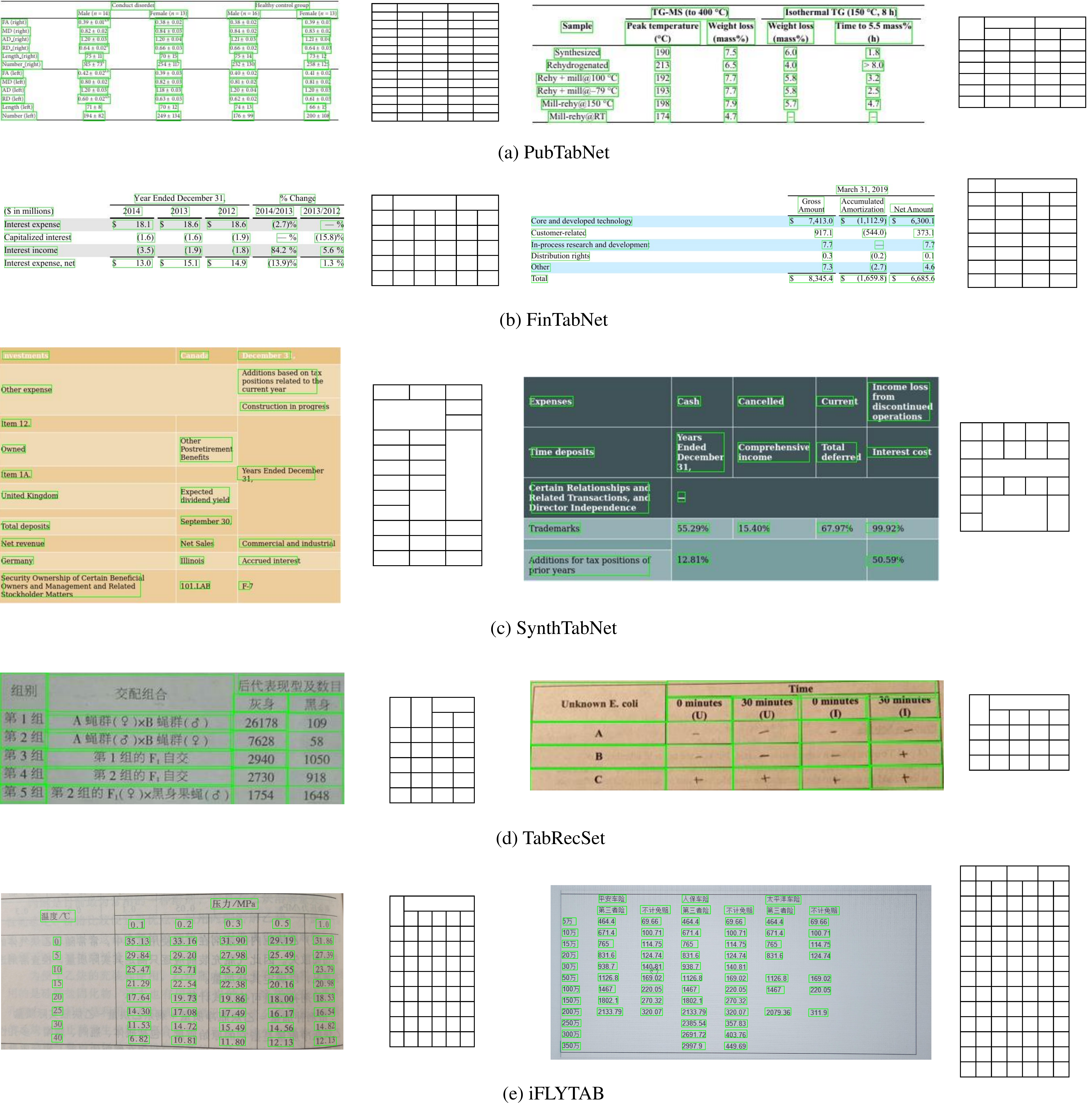,width=12cm}}
\caption{Visualization effects of BGTR. The predicted cell bounding boxes are depicted using \textcolor{green}{green} polygonal frames. From top to bottom, the sequence is PubTabNet~\cite{EDD}, FinTabNet~\cite{GTE}, SynthTabNet~\cite{Tableformer}, TabRecSet~\cite{TabRecSet} and iFLYTAB~\cite{SEMV2}. The first and third columns show the visualization of bounding boxes, while the second and fourth columns display the visualization of HTML sequences.}
\label{fig:6}
\end{figure}

We illustrate some visualization of BGTR in PubTabNet~\cite{EDD}, FinTabNet~\cite{GTE}, SynthTabNet~\cite{Tableformer}, TabRecSet~\cite{TabRecSet} and iFLYTAB~\cite{SEMV2}. As shown in Fig.~\ref{fig:6}, BGTR is adept at handling a wide range of scenarios and complex table structures. This includes tables with row and column spans, those containing multi-line text, as well as instances with empty cells. Moreover, it demonstrates strong robustness in both digital documents and natural scene environments. 

\begin{table}[t]
\begin{center}
\tabcolsep=0.3cm
\renewcommand\arraystretch{1.0}
\caption{Ablation studies of module design. \textbf{BG} signifies bbox-guided HTML generation. \textbf{PT} signifies the progressive training method.}
\vspace{2mm}
\label{tab:3}
\begin{tabular}{cc|ccc}
\hline
\multicolumn{2}{c|}{Methods} & \multicolumn{3}{c}{TEDS-S} \\ \hline
 BG &  PT & Simple & Complex & All \\ \hline
   &  & 98.30 & 86.50 & 95.54 \\
  $\checkmark$ &  &  98.35 & 89.27 & 96.23  \\
   $\checkmark$ &  $\checkmark$ &  \textbf{98.65} & \textbf{92.47} & \textbf{97.21}  \\ \hline
\end{tabular}
\end{center}
\end{table}

\begin{table}[t]
\begin{center}
\tabcolsep=0.25cm
\renewcommand\arraystretch{1.0}
\caption{Ablation studies of advancement training stage training
method. \textbf{SD} indicates Share Decoder. \textbf{BD} indicates Bbox Decoder. \textbf{BGD} indicates Bbox-Guided Structure Decoder.}
\vspace{2mm}
\label{tab:4}
\begin{tabular}{ccc|ccc}
\hline
\multicolumn{3}{c|}{Methods} & \multicolumn{3}{c}{TEDS-S} \\ \hline
 SD & BD & BGD & Simple & Complex & All \\ \hline
 $\checkmark$ &    &  & 98.60 & 92.21 & 97.11 \\
 $\checkmark$ & $\checkmark$ &  & 98.56 & 91.96 & 97.02  \\
 $\checkmark$ &  & $\checkmark$  & \textbf{98.65} & \textbf{92.47} & \textbf{97.21}  \\
 $\checkmark$ & $\checkmark$ & $\checkmark$ & 98.60 & 92.18 & 97.10 \\ \hline
\end{tabular}

\vspace{-3mm}

\end{center}
\end{table}

\subsection{Ablation Studies}

For simplicity, we conduct ablation experiments on TabRecSet~\cite{TabRecSet}. Several experiments were conducted to validate the effectiveness of our methods. 

\begin{figure}[t]
\centerline{\epsfig{figure=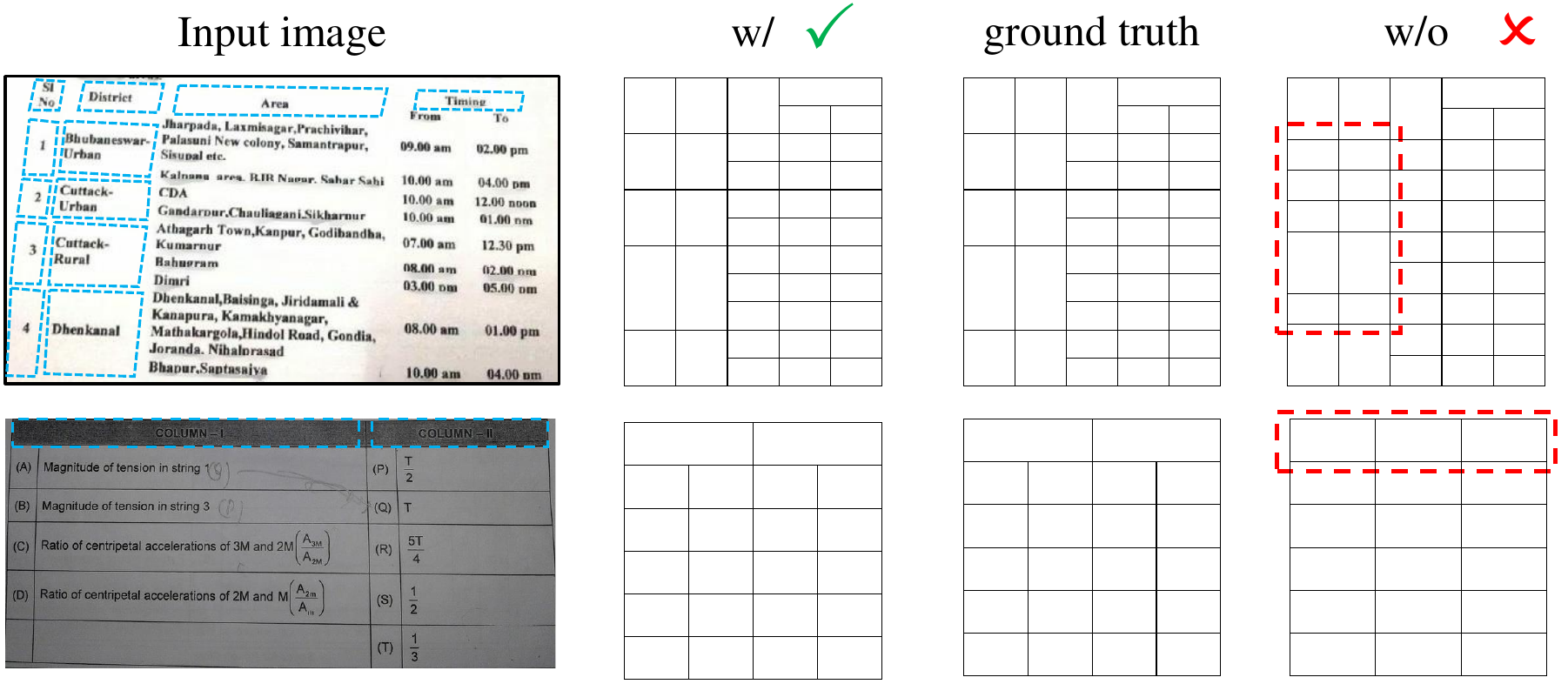,width=11cm}}
\caption{Comparison of HTML visualization w/ and w/o progressive training on TabRecSet~\cite{TabRecSet}, the \textcolor{red}{red} dotted boxes indicate error results, the \textcolor{cyan}{blue} dotted boxes indicate spanning cells.}
\vspace{-3mm}
\label{fig:4}
\end{figure}

\textbf{Effectiveness of module design}. As indicated in Table~\ref{tab:3}, \textbf{BG} signifies bbox-guided HTML generation. \textbf{PT} signifies the progressive training method, we constructed the baseline experiment following the previous methods~\cite{Tableformer,VAST,TableMaster,chen2022complex} which overlook the explicit utilization of bbox information when predicting HTML sequences. Utilizing \textbf{BG} significantly improves the TEDS-S score by 2.77\% on complex tables, indicating the effectiveness of guiding HTML sequence generation with bbox information in complex table scenes.  Meanwhile, \textbf{PT} enhances the model's generalization capabilities, particularly in handling complex tables, proving the effectiveness of the progressive training method. As shown in Fig.~\ref{fig:4}, using the progressive training method can yield better results on spanning cells and deformed tables.

\begin{table}[t]
\begin{center}
\tabcolsep=0.4cm
\renewcommand\arraystretch{1.0}
\caption{Ablation studies of ${\lambda }$ in loss function.}
\vspace{2mm}
\label{tab:5}
\begin{tabular}{c|ccc}
\hline
\multirow{2}{*}{${\lambda }$} & \multicolumn{3}{c}{TEDS-S} \\ \cline{2-4}
   & Simple & Complex & All \\ \hline
 0.5 & 98.49 & 91.82 & 96.94 \\
 1 & \textbf{98.65} & \textbf{92.47} & \textbf{97.21} \\
 2 & 98.57 & 91.92 & 97.02 \\ \hline

\end{tabular}
\vspace{-1mm}
\end{center}
\end{table}

\textbf{Effectiveness of advancement training stage training method}. As indicated in Table~\ref{tab:4}, placing a check mark ($\checkmark$) signifies that the module continues to the advancement training stage of training, building upon the foundation training stage, while its absence indicates starting the training anew. From the results, we can see that \textbf{SD} (Share Decoder) and \textbf{BGD} (Bbox-Guided Structure Decoder) are very helpful for the training in the advancement training stage. This indicates that training in the foundation training stage with a large amount of data enables the model to learn a wide variety of table structures. However, due to the data differences between two stages, \textbf{BD} (Bbox Decoder) is not much of a help for the training in the advancement training stage.

\textbf{Effectiveness of $\bm {\lambda }$ in loss function.} As indicated in Table~\ref{tab:5}, the table indicates that deep supervision positively impacts performance. However, the numerical results demonstrate a notable consistency across various trade-off parameter settings. For simplicity in model training, we recommend using ${\lambda }=1$ in practical applications.

\section{Conclusion}

In this paper, we introduced BGTR, a novel framework that explicitly use bbox information to guide the generation of HTML sequences. Besides, to alleviate the problem of insufficient data leading to inadequate training of bbox-guided HTML generation in natural scenes, we adopted a progressive training method for natural scene tables and introduced SNSTab, a large synthetic table dataset in natural scenes. Experimental results on five benchmark datasets demonstrate that the proposed method achieves state-of-the-art performance.

\subsubsection{Acknowledgements} The research is partially supported by National Key R\&D Program of China (2023YFC3502900), National Natural
Science Foundation of China (No. 62176093, 61673182), Key Realm R\&D Program of Guangzhou (No.
202206030001), Guangdong Provincial Science and Technology Plan (No. 2023A0505030016).

%
%
%
\bibliographystyle{splncs04}
\bibliography{ICPR2024template}
%




%
\clearpage

\appendix

\section{SNSTab}

\subsection{Dataset creation}

The creation of the SNSTab dataset is divided into three phases: table generation, table transformation, and background synthesis. In this section we will cover the details of each step.

{\bf Table generation.} When generating digital document tables, we set the maximum number of rows in the table to 20, the maximum number of columns to 15, the minimum number of rows to 2, and the minimum number of columns to 2. In addition, in order to get spanning cells, we will randomly merge the rows and columns of the cells, which will not exceed 40\% of the total number of cells. The text in the cells is mainly from some common words and phrases in Chinese and English, and the length of the text will not exceed 10. Since table image generation using browser rendering is time consuming, it takes about 20 hours on average to generate 10,000 table images.

{\bf Table transformation.} After the source points are selected, the source points move randomly to form the target points. The movement of the source points do not exceed 10\% of the width of the image horizontally and 10\% of the height of the image vertically. In fact, not all tables in the natural scene are inclined and rotated, and in order to better simulate this situation, 20\% of the tables are not transformed.

\begin{figure}
\centerline{\epsfig{figure=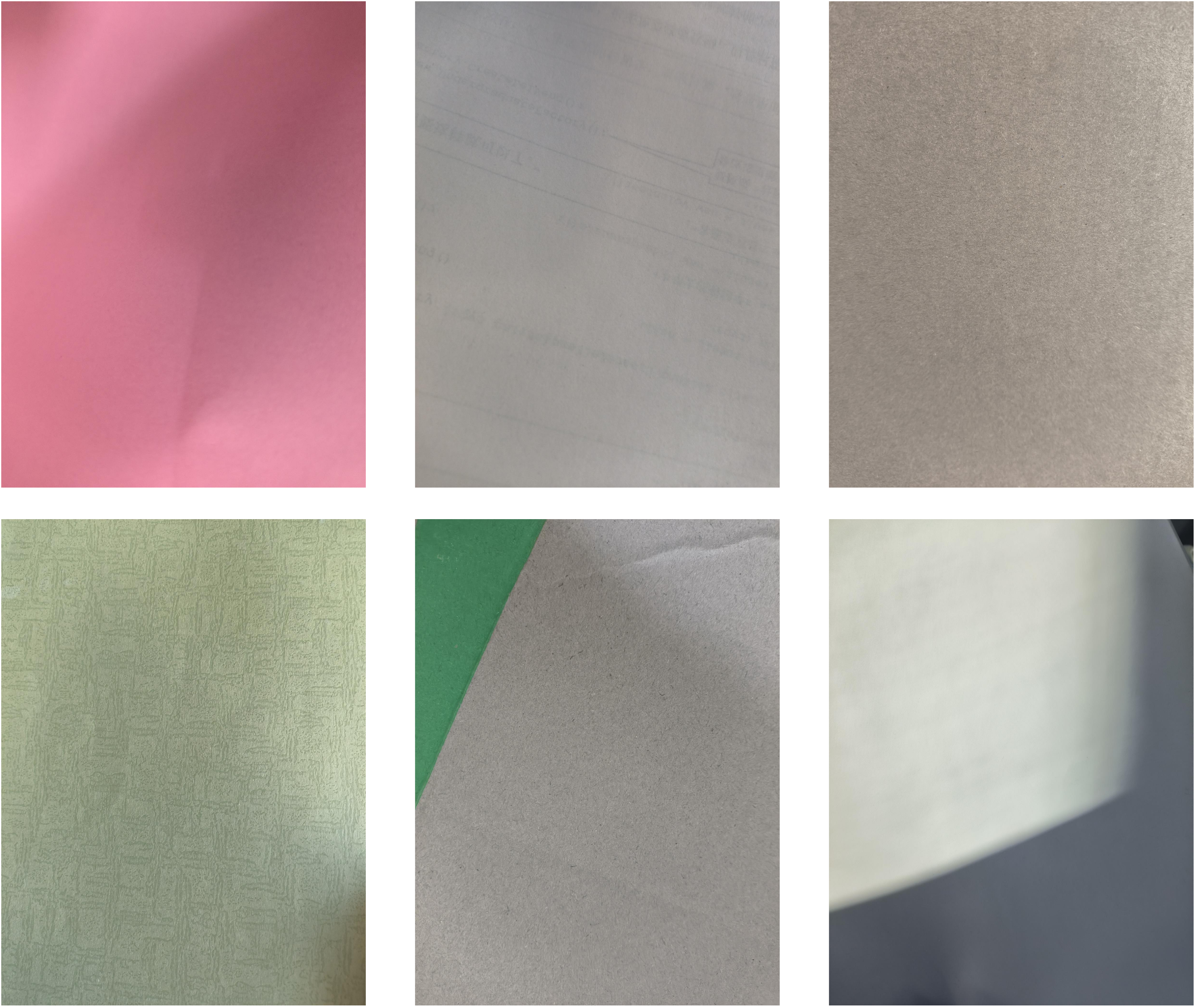,width=12cm}}
\caption{Samples in the background images. Including paper, walls and daily-life items.}
\label{fig:10}
\end{figure}

{\bf Background synthesis.} As shown in Fig.~\ref{fig:10}, in order to reduce the impact of the background image on the table text, we did not select a background with text when obtaining the background image. In the background synthesis, we will first obtain the mask region of the table image, then replace the mask region in the background image with the table image, and finally get the final sample.

\subsection{Samples}

\begin{figure}
\centerline{\epsfig{figure=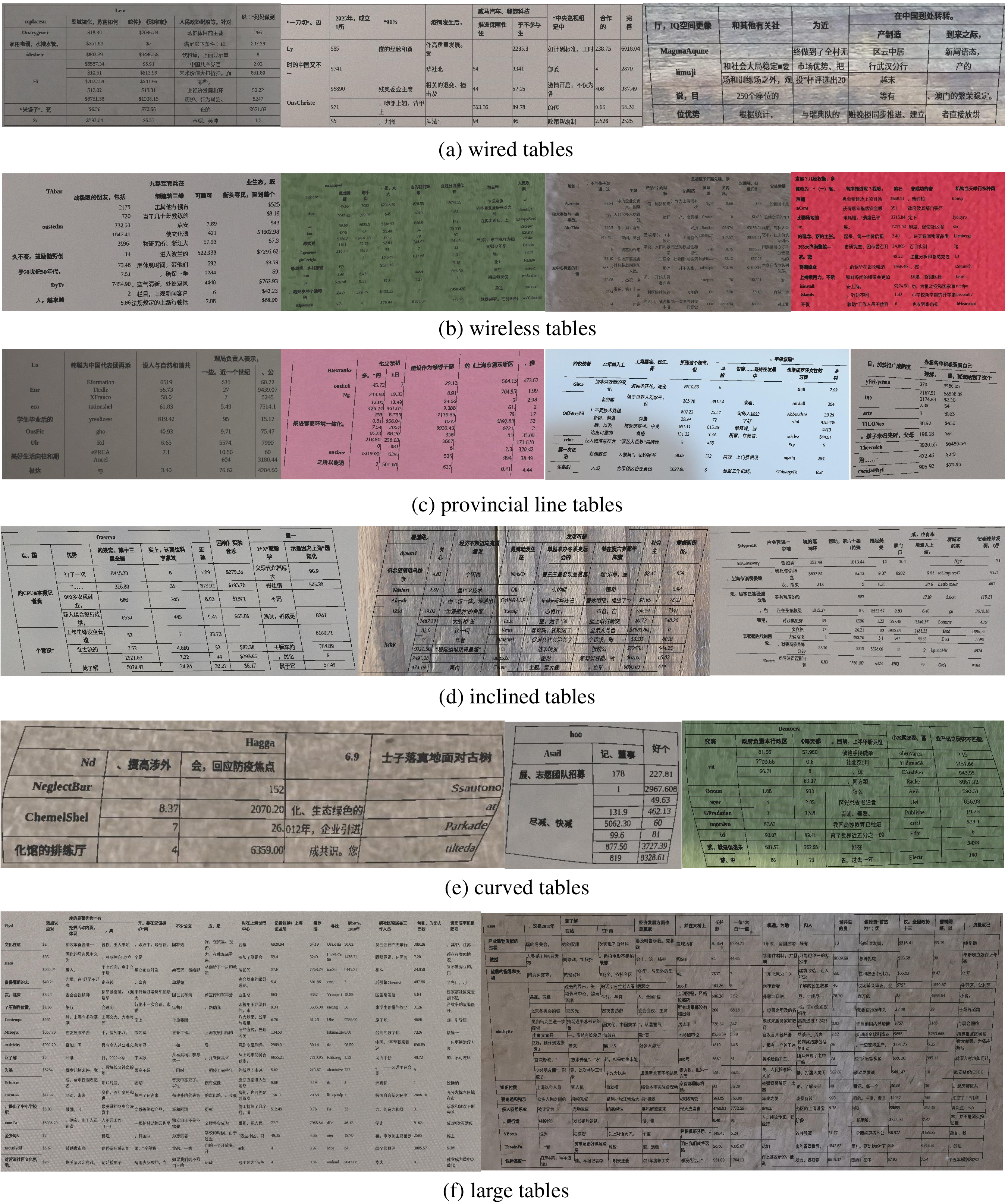,width=12cm}}
\caption{Samples in the SNSTab dataset. Including wired tables, wireless tables, provincial line tables, inclined tables, curved tables and large tables.}
\label{fig:5}
\end{figure}

SNSTab comprises a diverse range of tables within the dataset, including wired tables, Wireless tables, Provincial line tables, inclined tables, curved tables and large tables. Partial sample data is illustrated in Fig.~\ref{fig:5} for reference.

\subsection{Statistics}

\begin{figure}
\centerline{\epsfig{figure=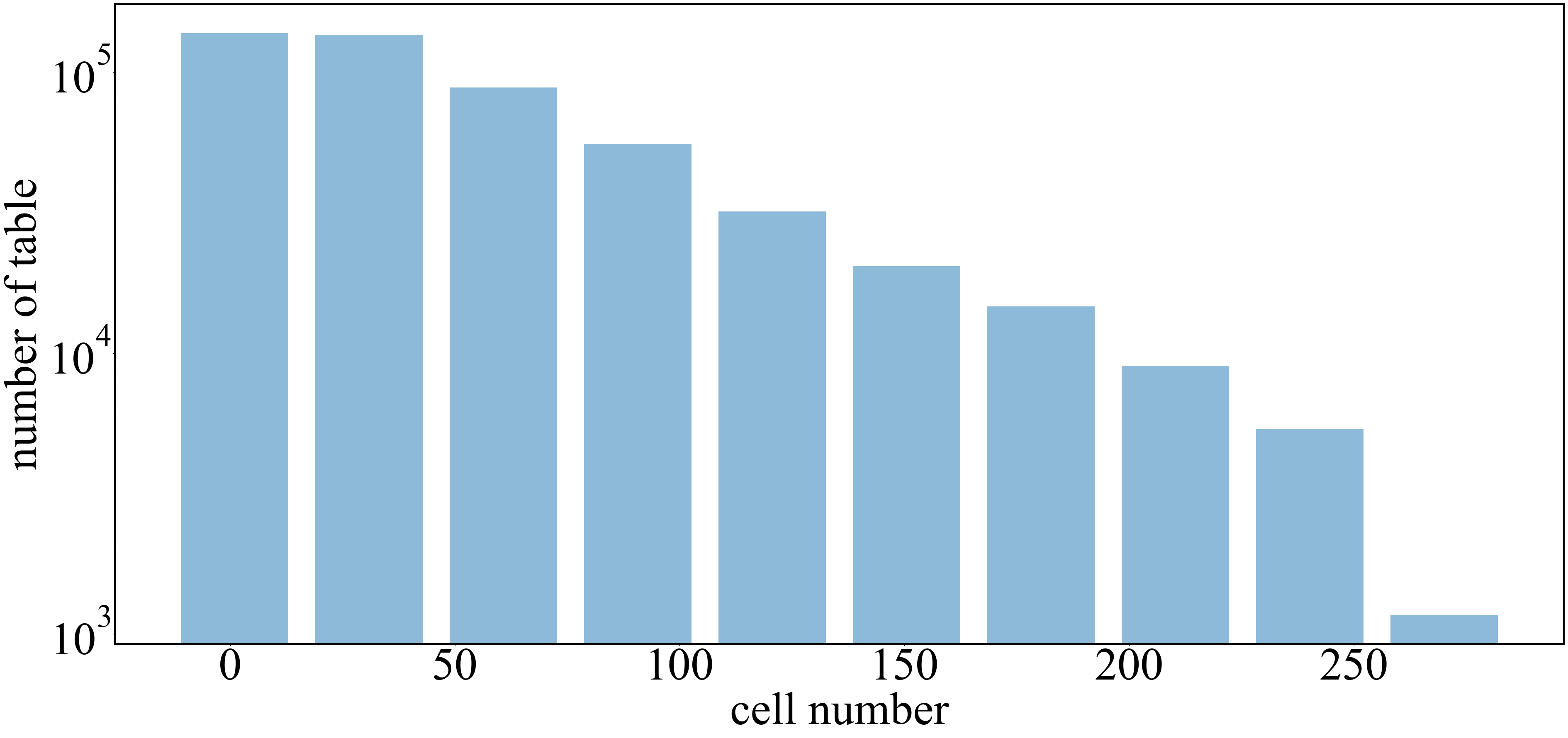,width=12cm}}
\caption{Statistics of cell number.}
\label{fig:cell number}
\end{figure}

To give a more complete picture of SNSTab, we have listed the following statistics:

{\bf Cell number}: this represents the number of cells contained in each table, as shown in Fig.~\ref{fig:cell number}

\begin{figure}
\centerline{\epsfig{figure=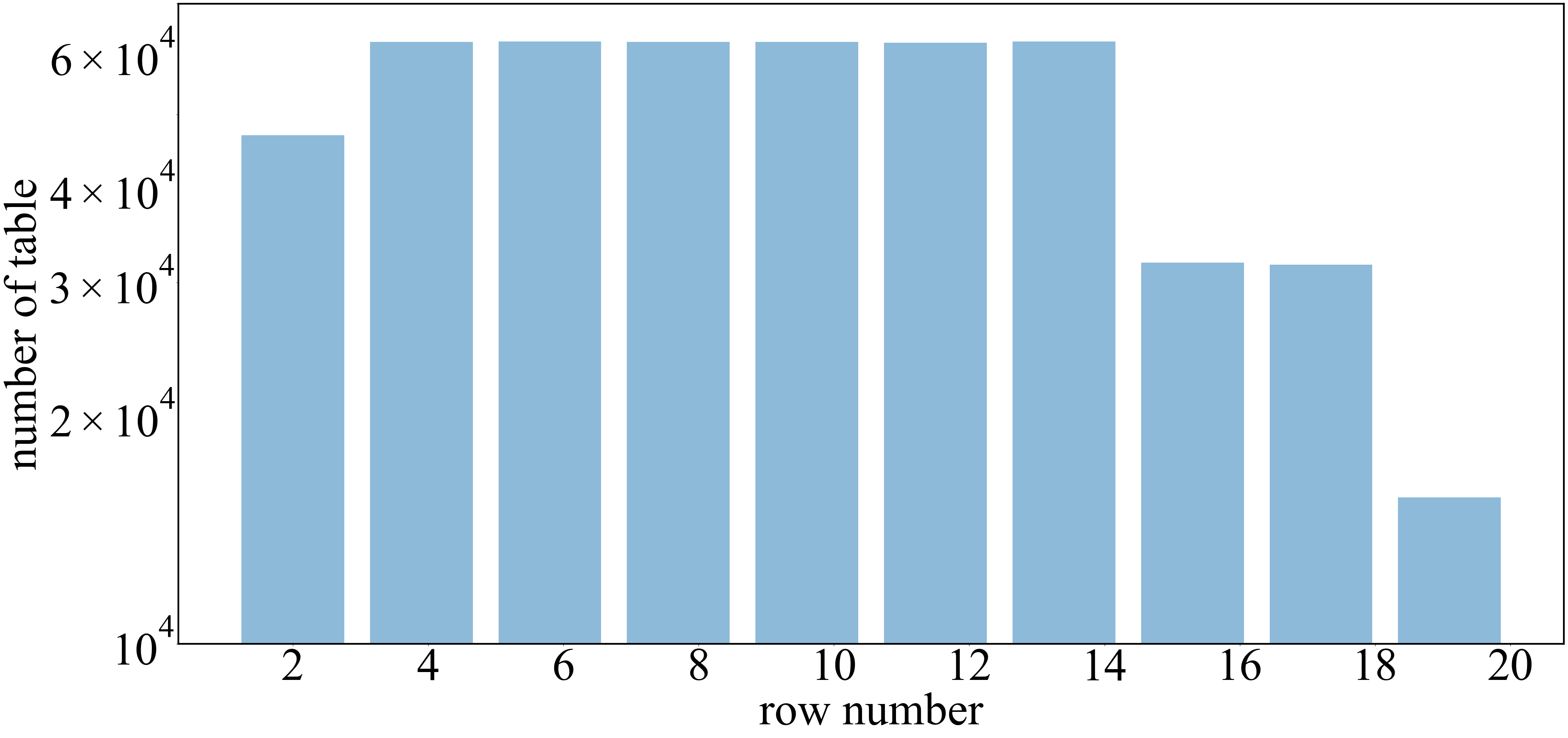,width=12cm}}
\caption{Statistics of row number.}
\label{fig:row number}
\end{figure}

{\bf Row number}: this represents the number of rows in each table, as shown in Fig.~\ref{fig:row number}

\begin{figure}
\centerline{\epsfig{figure=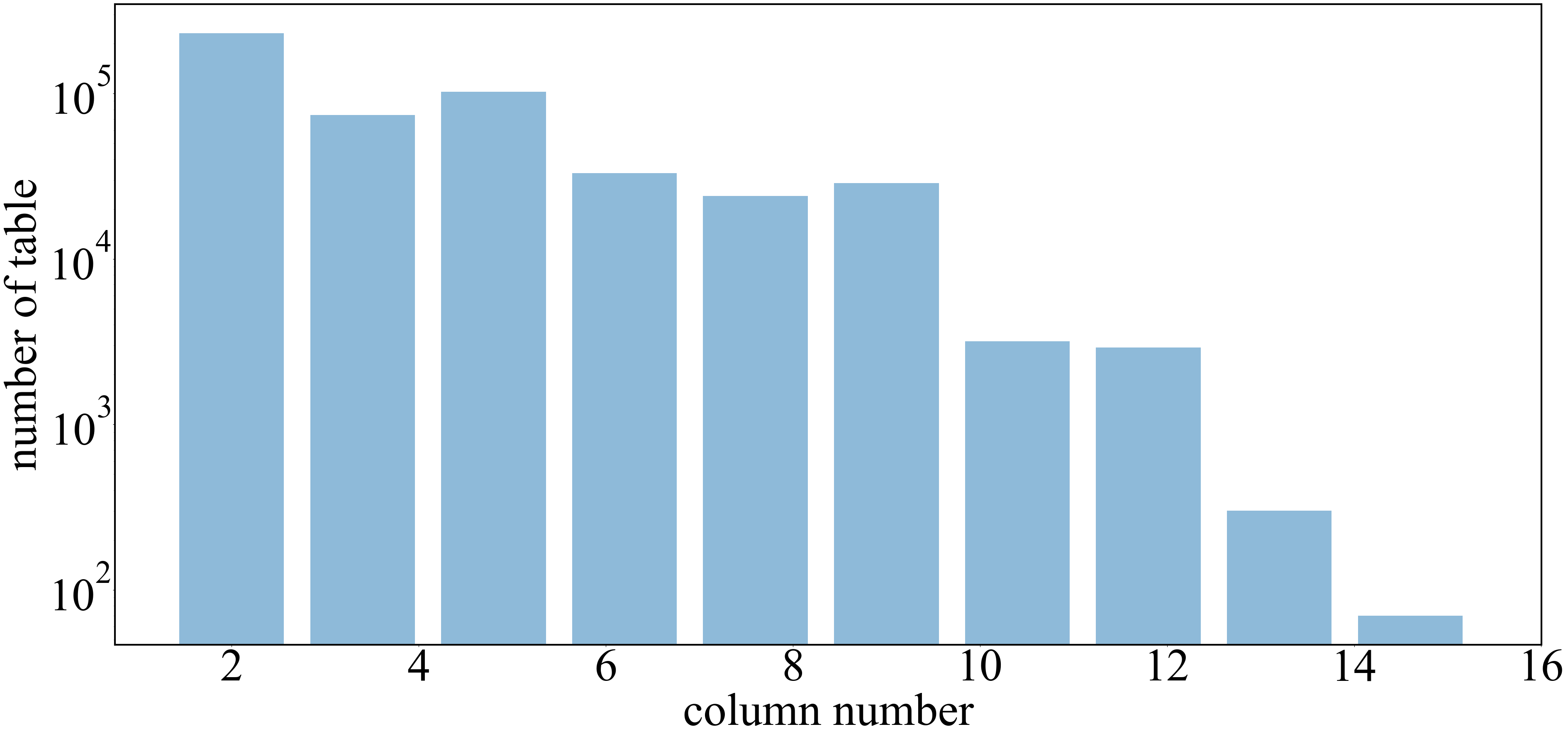,width=12cm}}
\caption{Statistics of column number.}
\label{fig:column number}
\end{figure}

{\bf Column number}: this represents the number of columns in each table, as shown in Fig.~\ref{fig:column number}

\begin{figure}
\centerline{\epsfig{figure=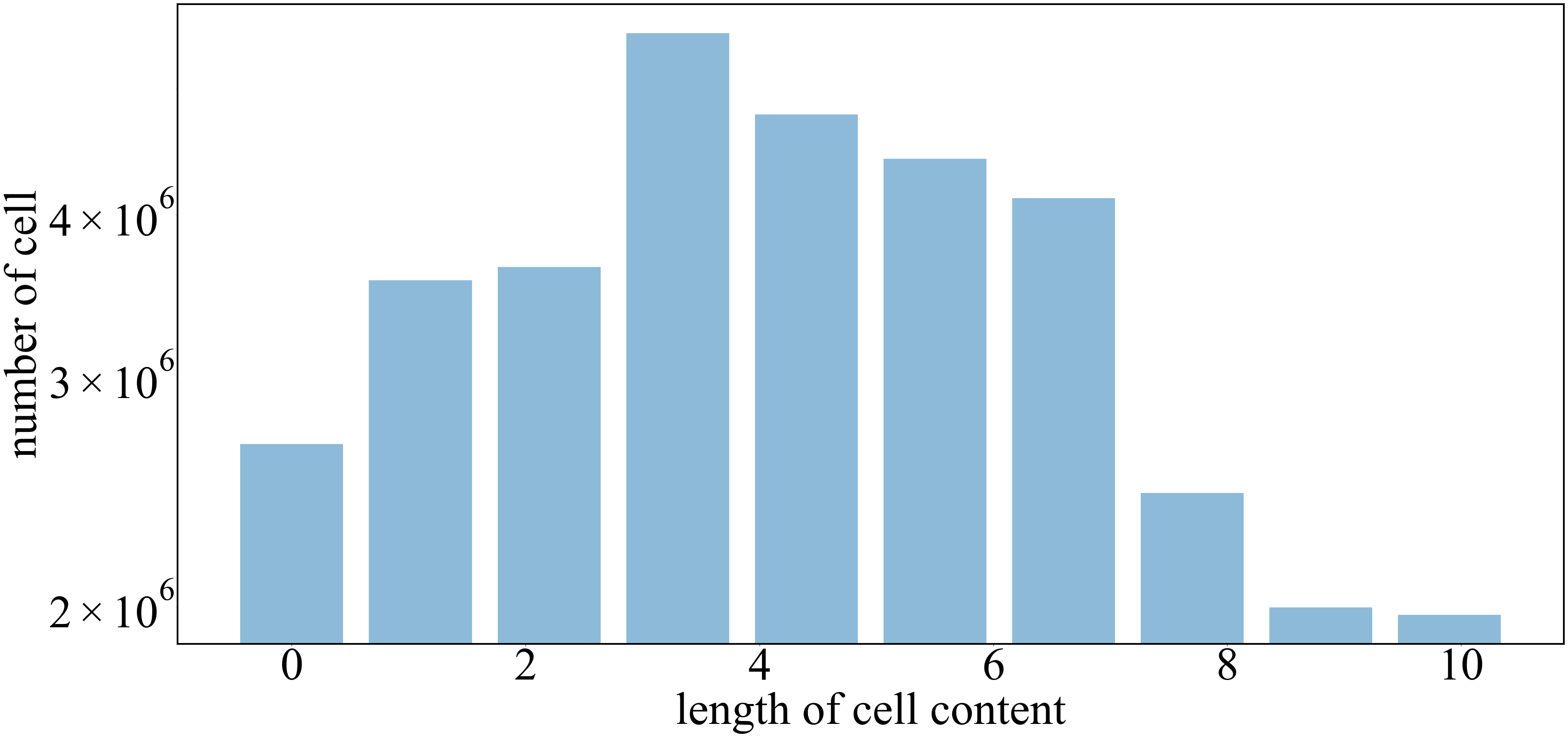,width=12cm}}
\caption{Statistics of length of cell content.}
\label{fig:content_len number}
\end{figure}

{\bf Length of cell content}: this represents the text length of each cell, as shown in Fig.~\ref{fig:content_len number}

\begin{figure}
\centerline{\epsfig{figure=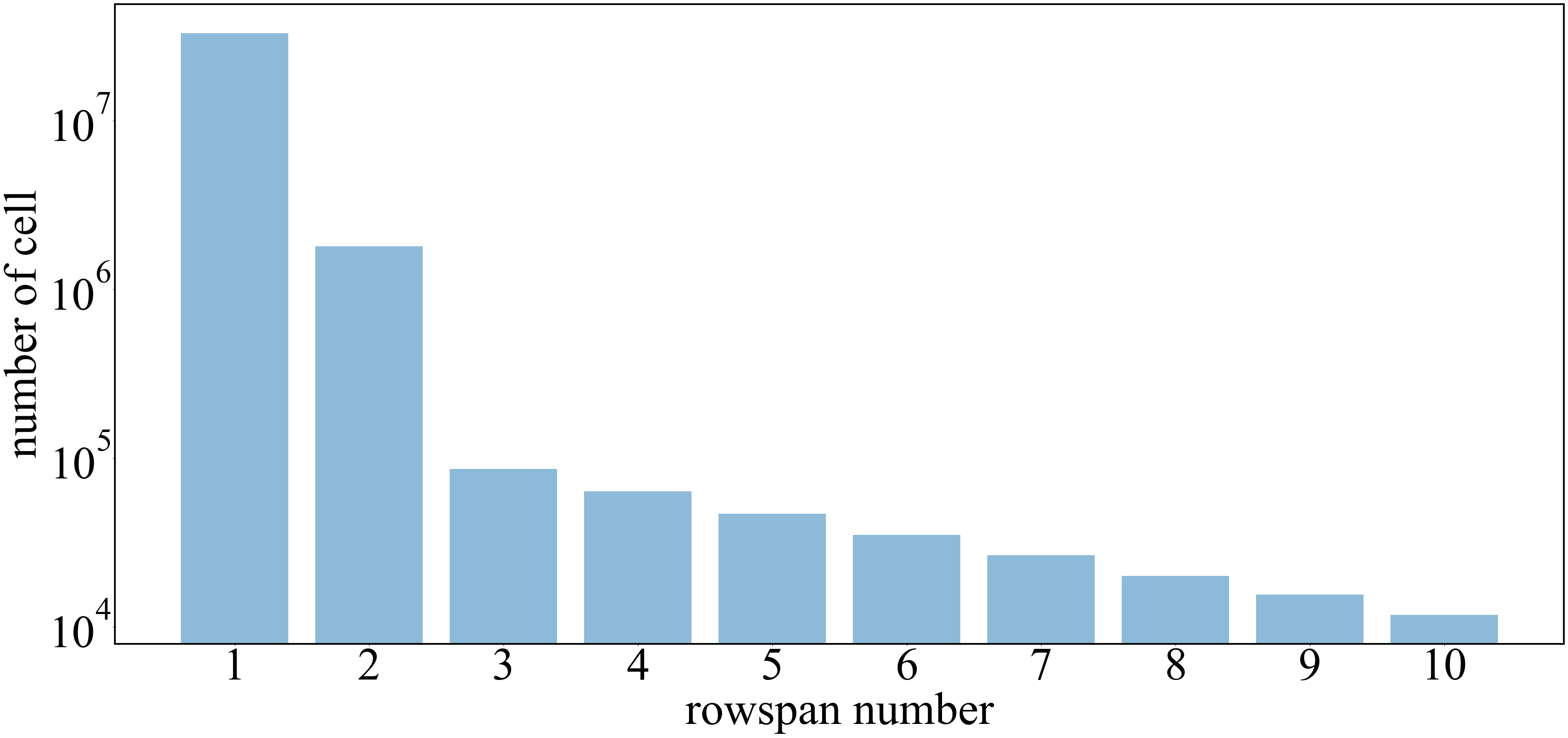,width=12cm}}
\caption{Statistics of rowspan number.}
\label{fig:rowspan number}
\end{figure}

{\bf Rowspan number}: this represents the number of rows that each cell spans, as shown in Fig.~\ref{fig:rowspan number}

\begin{figure}
\centerline{\epsfig{figure=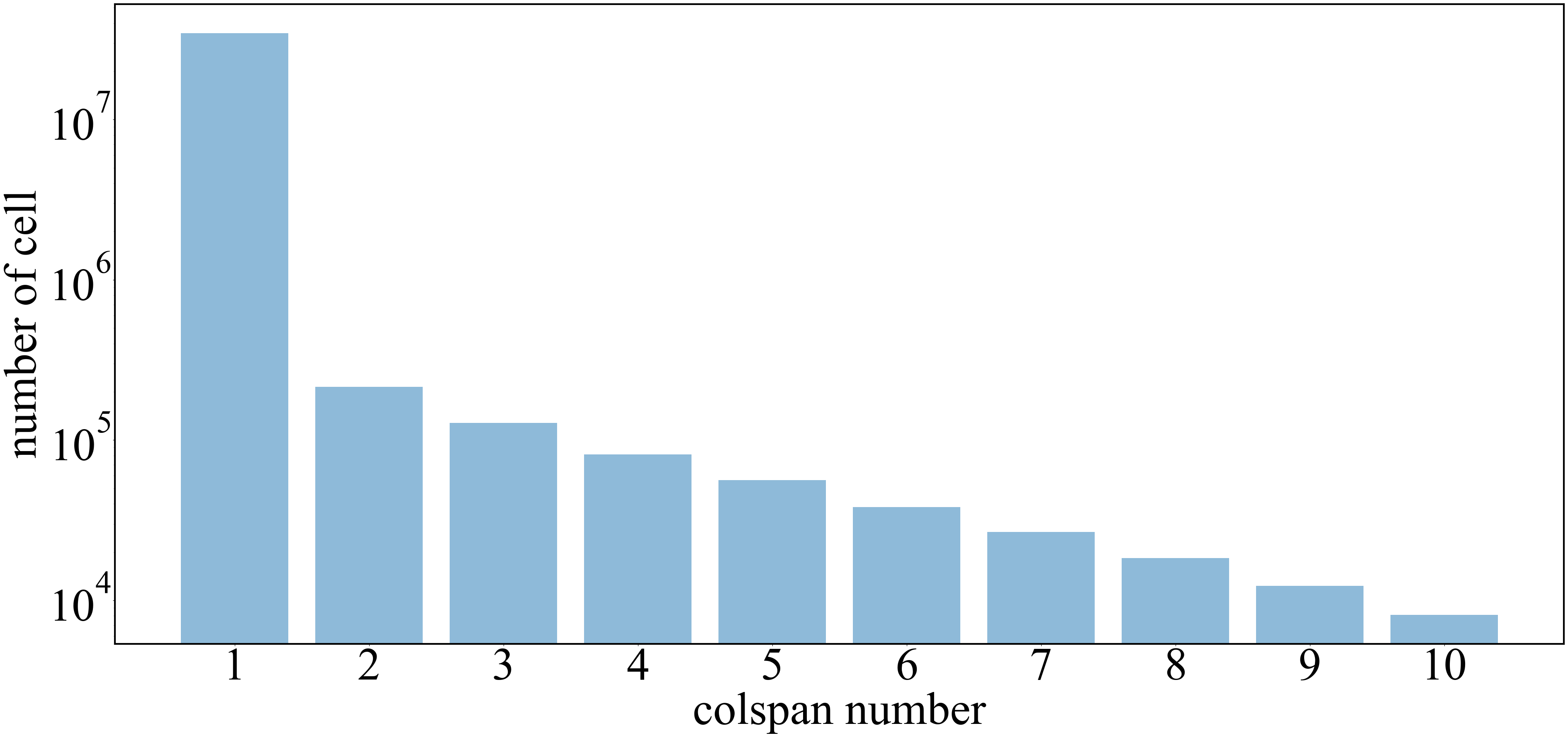,width=12cm}}
\caption{Statistics of colspan number.}
\label{fig:colspan number}
\end{figure}

{\bf Colspan number}: this represents the number of columns that each cell spans, as shown in Fig.~\ref{fig:colspan number}

\end{document}